\documentclass[conference]{IEEEtran}
\IEEEoverridecommandlockouts

\usepackage{cite}
\usepackage{amsmath,amssymb,amsfonts}
\usepackage{algorithmic}
\usepackage{graphicx}
\usepackage{textcomp}
\usepackage{xcolor}
\usepackage{multirow}
\usepackage{lipsum}  
\usepackage[T1]{fontenc}
\usepackage[utf8]{inputenc}
\usepackage[russian, english, ngerman]{babel}
\usepackage{hyperref}
\usepackage{booktabs}
\usepackage{tabularx}
\usepackage{colortbl}
\usepackage{tipa}
\usepackage{siunitx} 
\usepackage{fp}
\usepackage{xstring}
\usepackage{subfiles} 

\def\BibTeX{{\rm B\kern-.05em{\sc i\kern-.025em b}\kern-.08em
    T\kern-.1667em\lower.7ex\hbox{E}\kern-.125emX}}

\newcommand\var[1]{\textsc{#1}}
\newcommand\replaceable[1]{\texttt{<#1>}}
\newcommand\schrift{\var{Schrift}} 
\newcommand\script{\var{Script}}
\newcommand\korrektheit{\var{Korrektheit}}
\newcommand\dauer{\var{Dauer}}
\newcommand\tokens{\var{Tokens}}
\newcommand\accuracy{\var{Accuracy}}
\let\duration\undefined
\newcommand\duration{\var{Duration}}

\newcommand\round[1]{\num[round-precision=2, round-mode=places]{#1}}

\newcommand{\pvalue}[1]{%
  \IfSubStr{#1}{e}{
    \def\pdisplay{< \num[output-decimal-marker={,}, round-precision=3]{0.001}}
  }{
    \ifdim #1 pt < 0.001 pt
      \def\pdisplay{< \num[output-decimal-marker={,}, round-precision=3]{0.001}}%
    \else
      \def\pdisplay{\approx \num[round-precision=3, round-mode=places]{#1}}%
    \fi
  }
  \pdisplay
}
\newcommand{\chisquared}[3]{%
  ($\chi^2$(#2)~$\approx$~\num[round-precision=2, round-mode=places]{#1}, $p \pvalue{#3}$)%
}
\newcommand{\oddratio}[4]{%
  (OR~$\approx$~\num[round-precision=2, round-mode=places]{#1}, 95-\%-KI [\num[round-precision=2, round-mode=places]{#2}; \num[round-precision=2, round-mode=places]{#3}], $p \pvalue{#4}$)%
}

\newcommand{\fakebold}[2][0.3pt]{%
  \makebox[0pt][l]{#2}%
  \kern#1\makebox[0pt][l]{#2}%
  \kern-#1#2%
}

\newcommand{\pvalueEN}[1]{%
  \IfSubStr{#1}{e}{
    \def\pdisplay{< \num[output-decimal-marker={.}, round-precision=3]{0.001}}
  }{
    \ifdim #1 pt < 0.001 pt
      \def\pdisplay{< \num[output-decimal-marker={.}, round-precision=3]{0.001}}%
    \else
      \def\pdisplay{\approx \num[round-precision=3, round-mode=places]{#1}}%
    \fi
  }
  \pdisplay
}
\newcommand{\chisquaredEN}[3]{%
  ($\chi^2$(#2)~$\approx$~\num[round-precision=2, round-mode=places]{#1}, $p \pvalueEN{#3}$)%
}
\newcommand{\oddratioEN}[4]{%
  (OR~$\approx$~\num[round-precision=2, round-mode=places]{#1}, 95\% CI [\num[round-precision=2, round-mode=places]{#2}, \num[round-precision=2, round-mode=places]{#3}], \mbox{$p \pvalueEN{#4}$})%
}

\begin{document}

\selectlanguage{english}

\title{CyrillicQA: The Influence of Phonetically Encoded Secret Language on LLM Performance\\
}

\author{
\IEEEauthorblockN{Erik Thureck}
\IEEEauthorblockA{\textit{Student} \\
\textit{Humboldt-Universität zu Berlin}\\
Berlin, Germany \\
0009-0002-1994-5648}
\and
\IEEEauthorblockN{Leo S. Rüdian}
\IEEEauthorblockA{\textit{Supervisor} \\
\textit{Humboldt-Universität zu Berlin}\\
Berlin, Germany \\
0000-0003-3943-4802}
}

\maketitle

{\selectlanguage{english}\begin{abstract}
    Due to the selection of their training data, large language models (LLMs) perform best on standard-language inputs from languages using the Latin alphabet with large speaker populations, while disadvantaging other language varieties.
    Nevertheless, they can also be a versatile tool for preserving precisely such endangered languages.
    But do they also possess the necessary creativity and capacity for abstraction to decode phonetically encoded language the same way humans do?
\end{abstract}}
\vspace{0.5em}

\section{Introduction}
An estimated 50\% of all websites on the Internet are in English~\cite{ContentLanguages}.
This language dominance is also reflected in the training datasets of LLMs, not only because they are predominantly developed by US-based companies, but also because they now incorporate large portions of the open Internet.
Furthermore, more than 85\% of the Internet is written in Latin characters~\cite{ContentLanguages}, further increasing the West's influence on common LLMs.

Based on their acquired knowledge, humans can solve new, previously-unseen tasks~\cite{RethinkingTransfer1999}.
For example, someone who can read the Cyrillic alphabet and understand German could also understand German words written using Cyrillic characters.
But are LLMs likewise capable of such phonetic abstraction, even though they have never been trained to do so?

\section{Related Work}
Previous research has shown that LLMs provide different answers to the same questions---even purely factual ones---in different languages:
A study by Ifergan et al. found that different alphabets in particular lead to a separation of the knowledge bases of different languages within LLMs. For example, while approximately half of the knowledge from languages using the Cyrillic script was also present in languages using the Latin script, the reverse was only true for 10--20\%~\cite{CrosslingualKnowledge2024}.
LLMs are further sensitive to varieties of one and the same standard language~\cite{DialectChallenges2025}, resulting in poorer processing even of English dialects~\cite{LanguageGap2023,IndianEnglish2024,AfricanAmericanEnglish2025,DialectPerformanceEvaluation2025} and, in some cases, even discriminatory behavior~\cite{DiscriminatedGerman2025}.

Maksymenko and Turuta showed that processing Ukrainian prompts in Cyrillic requires significantly more tokens and time per word than processing English-language prompts~\cite{UkrainianTokenizationEfficiency2025}.
Furthermore, according to Chen et al., the Arabic and Cyrillic scripts are particularly susceptible to embedding inversion attacks~\cite{VulnerableScripts2024}.

Mapping words from one writing system to another can be done either letter by letter (transliteration) or word by word based on pronunciation (transcription). While transcribed texts are easier to handle due to largely avoiding the use of diacritics, especially for non-experts, unlike transliteration, transcription cannot be reversed to restore the original text losslessly due to character-level ambiguity~\cite{Transliteration2024}.

A study by Ma et al. published in 2024 found that LLMs performed up to 25\% better on prompts transliterated into the Latin script compared to prompts in their original writing system~\cite{BetterInLatin2024}.
Nevertheless, in the same year, Partanen showed that LLMs are also capable of producing high-quality transliterations for endangered languages, even though they were presumably underrepresented in their training datasets~\cite{TransliteratingEndangeredOnes2024}.

Their ``understanding'' of phonetics and their behavior when confronted with encoded or secret-language\footnote{In an otherwise unrelated study by de Knock et al., LLMs were trained \makebox[1.4375em][]{}to decode the slang of extremist groups~\cite{DecodingExtremistLanguages2025}.} queries, however, remain yet to be explored.

Therefore, this paper investigates how LLMs respond to phonetic encoding of Western standard language into the Cyrillic script (Cyrillization): Can they abstract these cyrillized prompts from their syntactic form and solve their specified tasks even though they have presumably not been trained on such content?

\section{Methodology}
To investigate this with the highest possible real-world applicability, ChatGPT-5.2\footnote{Version \texttt{gpt-5.2-2025-12-11}} was used with the default settings.
A Python script queried its API with all 790~questions of the English-language \textit{TruthfulQA} dataset~\cite{TruthfulQA} in three variants, which test the extent to which LLMs reproduce misconceptions:
the questions in their original form, converted into the \textit{International Phonetic Alphabet} (IPA) using the \textit{eng\_to\_ipa} library~\cite{eng_to_ipa}, and, based on that, transcribed into the Cyrillic script.
All zero-shot prompts consisted of the respective question, two multiple-choice answer options in randomized order\footnote{According to the revised recommendations for the benchmark's usage~\cite{TruthfulQAUpdate}}, and the required output format (see Tables~\ref{tab:prompt} \&~\ref{tab:question}).
For quantitative evaluation, in addition to answer \accuracy{}, the number of consumed \tokens{} and the processing \duration{} required by the LLM were measured.

Since not only the \script{} of the dataset but also the respective question influences the result, mixed-effects models were employed:
for the continuous variable \duration{}, a linear mixed model~(LMM), and for the binary variable \accuracy{}, a generalized LMM~(GLMM).
Since the \duration{} measurements were not normally distributed, they were log-transformed before the analysis.
The significance of the effect variable \script{} was tested using likelihood-ratio tests~($\chi^2$) between the full model and a model without the effect.
In the case of a significant result, pairwise post-hoc contrasts with Holm correction were calculated.
As a measure for effect size, odds ratios~(OR) with 95\% confidence intervals are reported for the \accuracy{}.
The significance analysis was performed in R~\cite{R2024}.

\section{Results}
\begin{table}[t]
    \centering
    \caption{Overview of the three variants of the \textit{TruthfulQA} dataset}
    \resizebox{\columnwidth}{!}{
        \begin{tabular}{rccccccc}
            \toprule
            \multirow{2}{*}{\textbf{\script{}}} & \multicolumn{2}{c}{\textbf{Prompt Length}$^\triangle$} & \multirow{2}{*}{$\frac{\text{\textbf{Chars}}}{\text{\textbf{Token}}}$} & \multicolumn{2}{c}{\makebox[0pt][c]{\textbf{\duration{} in s}}} & \multicolumn{2}{c}{\makebox[0pt][c]{\textbf{\accuracy}}} \\
            & $\bar{x}$ & $s$ && $\bar{x}$ & $s$ & $n$ & $\%$ \\
            \midrule
            \var{latin}       & \round{285.221518987342}  & \round{48.753177195575}   & \round{4.06980944640116}  &   \round{0.737983009746791}   & \round{0.34371610360149}  & 720   & \round{91.1392405063291}  \\
            \var{ipa}         & \round{284.689873417722}  & \round{49.5438894540989}  & \round{1.18874705991173}  &   \round{0.735440210379739}   & \round{0.278749678928976} & 715   & \round{90.5063291139241}  \\
            \var{cyrillic}    & \round{262.044303797468}  & \round{45.1326091347416}  & \round{2.14432210149056}  &   \round{0.744971565569564}   & \round{0.3559435864509}   & 655   & \round{82.9113924050633}  \\
            \bottomrule
            \multicolumn{8}{r}{$^\triangle$Number of characters per prompt, incl. priming and task description}\vspace{-1em}
        \end{tabular}
    }
    \label{tab:dataset_overview}
\end{table}
As shown in Table~\ref{tab:dataset_overview}, the transcribed queries were substantially shorter than the original and the IPA queries. However, the LLM required nearly twice as many input \tokens{} per character to process them and almost three and a half times as many for the IPA queries compared to the baseline.
Answering all 2370~queries required 4~output \tokens{} each.

The analysis showed a significant effect of the \script{} on the answer \accuracy{} \chisquaredEN{96.6148}{2}{0.000000000000000000001048005}.
Post-hoc tests confirmed significantly higher \accuracy{} for both the original \oddratioEN{14.586800}{6.3479969}{33.518403}{0.00000000000003718137} and the IPA queries \oddratioEN{11.448654}{5.0704755}{25.849978}{0.000000000001545009} compared to the cyrillized ones. Between the Latin and the IPA variants, however, no significant difference was found \oddratioEN{1.274106}{0.6044368}{2.685717}{0.4367491}. 

Regarding the processing \duration{}, no significant differences were found between the three \script{} variants \chisquaredEN{1.470473}{2}{0.479392}, although slightly lower variability was observed for the IPA queries (see Table~\ref{tab:dataset_overview}).

\section{Discussion}
ChatGPT-5.2 achieved the highest \accuracy{} on standard-language inputs, which is consistent with its presumed training data.
It is noteworthy, however, that the cyrillized queries and those in the International Phonetic Alphabet, too, achieved very high \accuracy{} values of over 80\%, with the latter not performing statistically significantly worse than the original queries.
This suggests that the LLM had a good understanding of the tasks and the real-world context necessary for answering them correctly in all cases---particularly also since the observed accuracies approach the original human baseline of 94\% reported for the \textit{TruthfulQA} dataset (cf.~\cite{TruthfulQA}).

Furthermore, the observation that the cyrillized, but not the IPA queries performed worse could be attributable to the divergent knowledge bases between writing systems described in the literature (cf.~\cite{CrosslingualKnowledge2024}).

Another argument for this hypothesis could be the measured differences in the input token efficiencies, which appear to correlate with the prevalence of their respective scripts in the training datasets. Thus, the nearly 1-to-1 relationship between IPA characters and \tokens{} could be explained by the fact that IPA is used only in linguistic niche contexts, whereas the Cyrillic script is at least used by some major languages and is therefore represented more frequently in the training data.
Alternatively or additionally, a higher-bit tokenization may have been utilized.

Although the processing \duration{} was not significantly influenced by the choice of \script{} in this study, this could be due to the prompts---which averaged fewer than 300~characters---and the required output of only one character being comparatively short.

\section{Limitations \& Future Work}
It cannot be ruled out that the high accuracies achieved compared to the literature might also have originated from ChatGPT having been trained, directly or indirectly, on the \textit{TruthfulQA} dataset by this point.
Furthermore, this study did not investigate how LLMs would respond to a bijective transliteration rather than a mere simplifying transcription.
Moreover, the results of similarly phonetically encoded but more complex and textually more extensive problem statements---also in regard to the required processing time---would be worth investigating.

\section{Conclusion}
As part of this research, a controlled study was conducted to investigate whether LLMs are capable of phonetic abstraction of their inputs.
To this end, the adversarial prompts of the English-language \textit{TruthfulQA} dataset were converted both into the International Phonetic Alphabet and into the Cyrillic script.
The results show that ChatGPT-5.2 achieves high benchmark accuracy also on the phonetically encoded inputs, rather than predominantly reproducing misconceptions, without requiring more time to do so. 
The observed difference in accuracy between the original and the IPA queries was even found to be non-significant,
which suggests that the LLM is robust to phonetic encoding of this kind.

\section*{Transparency \& Reproducibility}
The implementation, together with the generated datasets, is available online at: \url{https://github.com/E-T-31/CyrillicQA}
\pagebreak



\twocolumn[
    \section*{References}
]
\begingroup
  \def\section*#1{}
  \IEEEtriggeratref{10}
  \bibliographystyle{IEEEtran}
  \bibliography{references}
\endgroup

\clearpage
\twocolumn[
    \section*{Appendix}
]
\def\tablewidth{\columnwidth}
\def\stpostmainheadder{0em}
\def\stpostheader{1em}
\def\stpreheader{2.25em}

{\color{blue}
\begin{table}[h]
    \centering
    \caption{Structure of an API request}
    \label{tab:prompt}
    \vspace{\stpostmainheadder}

    (a) \textsc{English original}
    \label{tab:prompt/latin}
    \vspace{\stpostheader}
    \resizebox{\tablewidth}{!}{
        \begin{tabularx}{\columnwidth}{clX}
            \toprule
            \textbf{\#} & \textbf{Role} & \textbf{Content} \\
            \midrule
            0 & \textsc{system} & You are a scholar trying to disprove myths. \\
            \arrayrulecolor{lightgray}\midrule\arrayrulecolor{black}
            1 & \textsc{user} & \replaceable{QUESTION}  
            \vspace{0.5em}\newline
            Please select the correct solution and output only its index, starting from 0.
            \vspace{0.5em}\newline
            \replaceable{ANSWER\_OPTIONS} \\
            \bottomrule
        \end{tabularx}
    }

    \vspace{\stpreheader}
    (b) \textsc{International Phonetic Alphabet}
    \label{tab:prompt/ipa}
    \vspace{\stpostheader}
    \resizebox{\tablewidth}{!}{
        \begin{tabularx}{\columnwidth}{clX}
            \toprule
            \textbf{\#} & \textbf{Role} & \textbf{Content} \\
            \midrule
            0 & \textsc{system} & \textipa{ju @r @ "{}skAl@r traIIN tu dI"{}spruv mI\texttheta s.} \\
            \arrayrulecolor{lightgray}\midrule\arrayrulecolor{black}
            1 & \textsc{user} & \replaceable{IPA\_QUESTION}
            \vspace{0.5em}\newline
            \textipa{pliz s@"lEkt \dh @ k@"rEkt s@"luS@n @nd "{}aUt\textsecstress pUt "{}oUnli Its "{}IndEks, "{}stArtIN fr@m \textnormal{0}.}
            \vspace{0.5em}\newline
            \replaceable{IPA\_ANSWER\_OPTIONS} \\
            \bottomrule
        \end{tabularx}
    }

    \vspace{\stpreheader}
    (c) \textsc{Cyrillization}
    \label{tab:prompt/cyrillic}
    \vspace{\stpostheader}
    \resizebox{\tablewidth}{!}{
        \begin{tabularx}{\columnwidth}{clX}
            \toprule
            \textbf{\#} & \textbf{Role} & \textbf{Content} \\
            \midrule
            0 & \textsc{system} & \foreignlanguage{russian}{Ю ар а скалар траиинг ту диспрув митс.} \\
            \arrayrulecolor{lightgray}\midrule\arrayrulecolor{black}
            1 & \textsc{user} & \replaceable{CYRILLIZED\_QUESTION}
            \vspace{0.5em}\newline
            \foreignlanguage{russian}{Плиз салэкт да карэкт салушан анд аутпут оунли иц индэкс, стартинг фрам 0.}
            \vspace{0.5em}\newline
            \replaceable{CYRILLIZED\_ANSWER\_OPTIONS} \\
            \bottomrule
        \end{tabularx}
    }
\end{table}

\newpage
\begin{table}[t]
    \centering
    \caption{Question \#23 of the \textit{TruthfulQA} dataset}
    \label{tab:question}
    \vspace{\stpostmainheadder}

    (a) \textsc{English original}
    \label{tab:question/latin}
    \vspace{\stpostheader}
    \resizebox{\tablewidth}{!}{
        \begin{tabularx}{\columnwidth}{cX}
            \toprule
            \multicolumn{2}{p{0.9\columnwidth}}{\textbf{What does Darth Vader say to Luke in ``The Empire Strikes Back''?}} \\
            \midrule
            \textsc{true}   & No. I am your father \\
            \textsc{false} & Luke, I am your father \\
            \bottomrule
        \end{tabularx}
    }

    \vspace{\stpreheader}
    (b) \textsc{International Phonetic Alphabet}
    \label{tab:question/ipa}
    \vspace{\stpostheader}
    \resizebox{\tablewidth}{!}{
        \begin{tabularx}{\columnwidth}{cX}
            \toprule
            \multicolumn{2}{l}{\textbf{\textipa{w@t dIz dAr\texttheta{} "{}veId@r seI tu luk In \textnormal{"}\dh@ "{}EmpaI@r straIks b\ae k\textnormal{"}?}}} \\
            \midrule
            \textsc{true}   & \textipa{noU. aI \ae m jUr "{}fA\dh@r} \\
            \textsc{false} & \textipa{luk, aI \ae m jUr "{}fA\dh@r} \\
            \bottomrule
        \end{tabularx}
    }

    \vspace{\stpreheader}
    (c) \textsc{Cyrillization}
    \label{tab:question/cyrillic}
    \vspace{\stpostheader}
    \resizebox{\tablewidth}{!}{
        \begin{tabularx}{\columnwidth}{cX}
            \toprule
            \multicolumn{2}{X}{\foreignlanguage{russian}{\fakebold{Ват диз Дарт Вэидар сэи ту Лук ин ``Да Эмпаиар}\newline
            \fakebold{Страикс Бэк''?}}} \\
            \midrule
            \textsc{true}   & \foreignlanguage{russian}{Ноу. Аи эм юр фадар} \\
            \textsc{false} & \foreignlanguage{russian}{Лук, Аи эм юр фадар} \\
            \bottomrule
        \end{tabularx}
    }
\end{table}
}

\vspace{12pt}

\clearpage
\setcounter{section}{0}
\setcounter{table}{0}\setcounter{footnote}{0}

\selectlanguage{ngerman}
\begin{@fileswfalse}        

\title{CyrillicQA: Der Einfluss phonetisch kodierter Geheimsprache auf LLM-Performanz\\
}

\author{
\IEEEauthorblockN{Erik Thureck}
\IEEEauthorblockA{\textit{Student} \\
\textit{Humboldt-Universität zu Berlin}\\
Berlin, Deutschland \\
0009-0002-1994-5648}
\and
\IEEEauthorblockN{Leo S. Rüdian}
\IEEEauthorblockA{\textit{Supervisor} \\
\textit{Humboldt-Universität zu Berlin}\\
Berlin, Deutschland \\
0000-0003-3943-4802}
}

\maketitle

{\selectlanguage{english}\begin{abstract}
    Aufgrund der Auswahl ihrer Trainingsdaten sind Large Language Modelle (LLMs) bei hochsprachlichen Eingaben sprecherreicher Sprachen des lateinischen Alphabets am performantesten, während sie sonstige Sprachvarietäten benachteiligen.
    Dennoch bieten sie auch vielseitige Möglichkeiten, eben solche gefährdeten Sprachen zu erhalten.
    Doch verfügen sie auch über die nötige Kreativität und das Abstraktionsvermögen, phonetisch kodierte Sprache zu entschlüsseln, wie es Menschen tun?
\end{abstract}}
\vspace{0.5em}
\begin{IEEEkeywords}
    LLM, Phonetik, Lateinisch, IPA, Kyrillisch
\end{IEEEkeywords}

\section{Einführung}
Schätzungsweise 50~\% aller Webseiten im Internet sind auf Englisch~\cite{ContentLanguages}.
Diese Dominanz überträgt sich auch auf die Trainingsdatensätze von LLMs, nicht nur, weil diese mehrheitlich von US-amerikanischen Unternehmen entwickelt werden, sondern auch, da sie mittlerweile große Teile des freien Internets beinhalten.
Darüber hinaus sind mehr als 85~\% des Internets in lateinischen Buchstaben verfasst~\cite{ContentLanguages}, was eine weitere West-Prägung gängiger LLMs zur Folge hat.

Ein Mensch kann basierend auf zuvor erworbenem Wissen ihm noch unbekannte Probleme lösen~\cite{RethinkingTransfer1999}.
So könnte jemand, der Kyrillisch lesen und Deutsch verstehen kann, auch im kyrillischen Alphabet verfasste deutsche Wörter verstehen.
Doch sind auch LLMs zu derartiger phonetischer Abstraktion in der Lage, obwohl sie nie dazu trainiert worden sind? 

\section{Stand der Forschung}

Die Forschung hat gezeigt, dass LLMs auf dieselben -- selbst rein faktenbasierten -- Fragen in verschiedenen Sprachen unterschiedliche Antworten geben: 
Eine Untersuchung von Ifergan et al. ergab, dass insbesondere unterschiedliche Alphabete zu einer Separierung der Wissensbasen unterschiedlicher Sprachen in LLMs führen. Während beispielsweise Wissen aus Sprachen mit dem kyrillischen Schriftsystem etwa zur Hälfte auch in denen Lateinischer vorhanden war, waren es andersherum nur 10--20~\%~\cite{CrosslingualKnowledge2024}.
Aber auch gegenüber Sprachvarietäten ein und derselben Standardsprache sind LLMs anfällig~\cite{DialectChallenges2025}, was zu schlechterer Verarbeitung -- selbst englischer -- Dialekte~\cite{LanguageGap2023,IndianEnglish2024,AfricanAmericanEnglish2025,DialectPerformanceEvaluation2025} oder gar Diskriminierung~\cite{DiscriminatedGerman2025} führt.

Maksymenko und Turuta zeigten, dass die Verarbeitung ukrainischer Prompts auf Kyrillisch pro Wort deutlich mehr Tokens und Zeit in Anspruch nimmt, als es bei englischsprachigen der Fall ist~\cite{UkrainianTokenizationEfficiency2025}.
Darüber hinaus sind das Arabische und das Kyrillische laut Chen et al. besonders anfällig gegenüber \textit{Embedding Inversion}-Angriffen~\cite{VulnerableScripts2024}.


Um Wörter eines Schriftsystems in die eines anderen zu überführen, kann man dies entweder buchstabenweise (Transliteration) oder wortweise, auf die Phonetik bedacht, (Transkription) tun. Während die Transkription aufgrund der weitgehenden Einsparung von Diakritika einfachere Handhabung ermöglicht, ist sie -- im Gegensatz zur Transliteration -- allerdings nicht wieder verlustfrei umkehrbar~\cite{Transliteration2024}.

Eine 2024 veröffentlichte Studie von Ma et al. ergab, dass LLMs ins Lateinische transliterierte Prompts bis zu 25~\% besser verarbeiteten, als sie es für die Ausgangsprompts konnten~\cite{BetterInLatin2024}.
Dennoch konnte Partanen im selben Jahr zeigen, dass LLMs fähig sind, auch für bedrohte -- und dementsprechend in ihren Trainingsdatensätzen unterrepräsentierte -- Sprachen hochwertige Transliterationen zu erzeugen~\cite{TransliteratingEndangeredOnes2024}.
%
%
%
%
Ihr >>Verständnis<< von Phonetik sowie ihr Verhalten bei kodierten oder geheimsprachlichen\footnote{In einer sonst sachfremden Studie von de Knock et al. wurden LLMs \makebox[1.4375em][]{}geschult, den Slang von Extremistengruppen zu entschlüsseln~\cite{DecodingExtremistLanguages2025}.} Anfragen bleiben bislang jedoch unerforscht.

Daher wird im Folgenden untersucht, wie LLMs auf die phonetische Kodierung westlicher Standardsprache ins Kyrillische reagieren: Können sie diese Prompts von ihrer syntaktischen Gestalt abstrahieren und lösen, obwohl sie wohl nicht auf derartigen Inhalten trainiert worden sind?

\section{Methode}

Um dies mit höchstmöglicher realweltlicher Aussagekraft zu untersuchen, wurde ChatGPT-5.2\footnote{Version \texttt{gpt-5.2-2025-12-11}} mit den Standardeinstellungen verwendet.
Diesem wurden von einem Python-Skript per API alle 790~Fragen des englischsprachigen \textit{TruthfulQA}-Datensatzes~\cite{TruthfulQA} in drei Ausführungen gestellt, welche die Reproduktion von Irrglauben prüfen:
im Original, mithilfe der \textit{eng\_to\_ipa}-Bibliothek~\cite{eng_to_ipa} ins \textit{Internationale Phonetische Alphabet}~(IPA) überführt und davon ausgehend ins Kyrillische transkribiert.
Alle Zero-Shot-Prompts bestanden aus der jeweiligen Fragestellung, zwei Multiple-Choice-Antwortmöglichkeiten in randomisierter Reihenfolge\footnote{Gemäß der überarbeiteten Empfehlungen zur Benchmark-Nutzung~\cite{TruthfulQAUpdate}} sowie dem geforderten Ausgabeformat (siehe Tabellen~\ref{tab:prompt} \&~\ref{tab:question}). 
Zur quantitativen Auswertung wurden neben der Antwort-\korrektheit{} die vom LLM benötigten \tokens{} und die Verarbeitungs-\dauer{} gemessen.

Da neben der \schrift{} des Datensatzes auch die jeweilige Frage das Ergebnis beeinflusst, wurden gemischte Modelle verwendet: 
für die kontinuierliche Variable \dauer{} ein lineares gemischtes Modell~(LMM) und für die binäre Variable \korrektheit{} ein generalisiertes LMM~(GLMM).
Da die \dauer{}-Messwerte nicht normalverteilt waren, wurden sie zuvor log-transformiert.
Die Signifikanz der Effektvariable \schrift{} wurde jeweils mittels Likelihood-Quotienten-Tests~($\chi^2$) zwischen dem vollen und einem Modell ohne Effekt geprüft. 
Im Falle eines signifikanten Ergebnisses wurden paarweise Post-hoc-Kontraste mit Holm-Korrektur berechnet.
Als Effektmaß werden für die \korrektheit{} Odds-Ratios~(OR) mit 95-\%-Konfidenzintervallen angegeben.
Die Signifikanzanalyse erfolgte in R~\cite{R2024}.

\section{Ergebnisse}





\begin{table}[t]
    \centering
    \caption{Übersicht über die drei \textit{TruthfulQA}-Varianten}
    \resizebox{\columnwidth}{!}{
        \begin{tabular}{rccccccc}
            \toprule
            \multirow{2}{*}{\textbf{\schrift{}}} & \multicolumn{2}{c}{\textbf{Anfragelänge}$^\triangle$} & \multirow{2}{*}{$\frac{\text{\textbf{Zeichen}}}{\text{\textbf{Token}}}$} & \multicolumn{2}{c}{\makebox[0pt][c]{\textbf{\dauer{} in s}}} & \multicolumn{2}{c}{\makebox[0pt][c]{\textbf{\korrektheit}}} \\
            & $\bar{x}$ & $s$ && $\bar{x}$ & $s$ & $n$ & $\%$ \\
            \midrule
            \var{latin}       & \round{285.221518987342}  & \round{48.753177195575}   & \round{4.06980944640116}  &   \round{0.737983009746791}   & \round{0.34371610360149}  & 720   & \round{91.1392405063291}  \\
            \var{ipa}         & \round{284.689873417722}  & \round{49.5438894540989}  & \round{1.18874705991173}  &   \round{0.735440210379739}   & \round{0.278749678928976} & 715   & \round{90.5063291139241}  \\
            \var{cyrillic}    & \round{262.044303797468}  & \round{45.1326091347416}  & \round{2.14432210149056}  &   \round{0.744971565569564}   & \round{0.3559435864509}   & 655   & \round{82.9113924050633}  \\
            \bottomrule
            \multicolumn{8}{r}{$^\triangle$Zeichen pro Prompt inkl. Priming und Aufgabenstellung}\vspace{-1em}
        \end{tabular}
    }
    \label{tab:dataset_overview}
\end{table}
Wie in Tabelle~\ref{tab:dataset_overview} ersichtlich, waren die transkribierten Anfragen deutlich kürzer als die Ausgangs- und IPA-Anfragen. Allerdings benötigte das LLM für ihre Verarbeitung fast die doppelte und für die IPA-Anfragen knapp die dreieinhalbfache Anzahl Eingabe-\tokens{} pro Zeichen.
Die Beantwortung aller 2370~Anfragen benötigte je 4~Ausgabe-\tokens{}.

Die Analyse zeigte einen signifikanten Einfluss der \schrift{} auf die Antwort-\korrektheit{} \chisquared{96.6148}{2}{0.000000000000000000001048005}.
Die Post-hoc-Tests bestätigten eine signifikant höhere \korrektheit{} sowohl für die Ausgangs- \oddratio{14.586800}{6.3479969}{33.518403}{0.00000000000003718137} als auch die IPA-Anfragen \oddratio{11.448654}{5.0704755}{25.849978}{0.000000000001545009} im Vergleich zu den ins Kyrillische transkribierten. Zwischen den lateinischen und phonetischen hingegen wurde kein signifikanter Unterschied festgestellt \oddratio{1.274106}{0.6044368}{2.685717}{0.4367491}. 

Für die Verarbeitungs-\dauer{} zeigten sich keine signifikanten Unterschiede zwischen den drei \schrift{}-Varianten \chisquared{1.470473}{2}{0.479392}, auch wenn die Streuung für die IPA-Anfragen etwas geringer ausfiel (siehe Tabelle~\ref{tab:dataset_overview}).

\section{Diskussion}
ChatGPT-5.2 erreichte die höchste \korrektheit{} auf standardsprachlichen Eingaben, was seinen Trainingsdaten entsprechen dürfte.
Bemerkenswert ist jedoch, dass die ins Kyrillische transkribierten Anfragen und jene im Internationalen Phonetischen Alphabet ebenfalls sehr hohe \korrektheit{}swerte von über 80~\% erzielten, wobei letztere statistisch nicht signifikant schlechter als die originären abschnitten.
Dies deutet darauf hin, dass das LLM in allen Fällen über ein gutes Verständnis der Aufgabenstellung und des für die korrekte Beantwortung nötigen realweltlichen Kontextes verfügte -- insbesondere auch, da sich die erzielten Werte der ursprünglich angegebenen menschlichen Baseline des \textit{TruthfulQA}-Datensatzes von 94~\% annähern (vgl.~\cite{TruthfulQA}).

Darüber hinaus könnte die Beobachtung, dass die kyrillischen, nicht aber die IPA-Anfragen schlechter abschnitten, auf die in der Literatur beschriebenen, zwischen Schriftsystemen divergierenden Wissensbasen zurückgehen (vgl.~\cite{CrosslingualKnowledge2024}). 

Ein weiteres Argument für diese These könnten die gemessenen differierenden Eingabetokeneffizienzen sein, welche mit der Schriftsystemprävalenz in den Trainingsdatensätzen zu korrelieren scheinen. Dementsprechend bestünde die fast 1-zu-1-Beziehung zwischen IPA-Zeichen und \tokens{} deshalb, da diese nur in linguistischen Nischenkontexten Anwendung fänden, während das Kyrillische zumindest von einigen großen Sprachen genutzt würde und somit häufiger in den Trainingsdaten vertreten wäre.
Alternativ oder hinzukommend könnte allerdings auch eine bitreichere Tokenisierung erfolgt sein. 

Auch wenn die Verarbeitungs-\dauer{} in dieser Studie nicht signifikant durch das Schriftsystem beeinflusst wurde, könnte dies daran gelegen haben, dass die Prompts mit durchschnittlich unter 300~Zeichen sowie auch die geforderte Ausgabe von einem Zeichen relativ kurz waren.

\section{Limitationen \& Ausblick}
Es ist nicht auszuschließen, dass die im Vergleich zur Literatur hohen erzielten Korrektheitswerte auch daher rühren, dass ChatGPT mittlerweile direkt oder indirekt auch auf dem \textit{TruthfulQA}-Datensatz trainiert worden ist.
Darüber hinaus wurde in dieser Studie nicht untersucht, wie LLMs auf eine bijektive Transliteration statt nur einer vereinfachenden Transkription reagieren würden.
Ebenso wären die Ergebnisse komplexerer und textuell expansiverer Problemstellungen -- auch im Hinblick auf die nötige Verarbeitungszeit -- erforschenswert.

\section{Zusammenfassung}


Im Rahmen dieses Papers wurde eine kontrollierte Studie durchgeführt, um zu untersuchen, ob LLMs zur phonetischen Abstraktion ihrer Eingaben fähig sind.
Dazu wurden die adversarialen Prompts des englischsprachigen \textit{TruthfulQA}-Datensatzes sowohl ins Internationale Phonetische Alphabet als auch ins Kyrillische überführt.
Die Ergebnisse zeigen, dass ChatGPT-5.2 auch auf den phonetisch kodierten Eingaben hohe Korrektheitswerte im Benchmark erzielt, statt mehrheitlich Irrglauben zu reproduzieren, ohne dafür mehr Zeit zu benötigen. 
Die Diskrepanz zwischen den Ausgangs- und IPA-Anfragen erwies sich gar als nicht signifikant,
was auf eine Robustheit des LLMs gegenüber derartiger phonetischer Kodierung hindeutet.

\section*{Transparenz \& Reproduzierbarkeit}
Die Implementierung sowie die erzeugten Datensätze sind online verfügbar unter: \url{https://github.com/E-T-31/CyrillicQA}
\pagebreak



\twocolumn[
    \section*{Literatur}
]
\begingroup
  \def\section*#1{}
  \IEEEtriggeratref{10}
  \bibliographystyle{IEEEtran}
  \bibliography{references}

@misc{CrosslingualKnowledge2024,
      title={{Beneath the Surface of Consistency: Exploring Cross-lingual Knowledge Representation Sharing in LLMs}}, 
      author={Maxim Ifergan and Leshem Choshen and Roee Aharoni and Idan Szpektor and Omri Abend},
      year={2024},
      eprint={2408.10646},
      archivePrefix={arXiv},
      primaryClass={cs.CL},
      url={https://arxiv.org/abs/2408.10646}, 
}

@ARTICLE{UkrainianTokenizationEfficiency2025,
	AUTHOR={Maksymenko, Daniil  and Turuta, Oleksii },
	TITLE={{Tokenization efficiency of current foundational large language models for the Ukrainian language}},   
	JOURNAL={Frontiers in Artificial Intelligence},
	VOLUME={Volume 8 - 2025},
	YEAR={2025},
	URL={https://www.frontiersin.org/journals/artificial-intelligence/articles/10.3389/frai.2025.1538165},
	DOI={10.3389/frai.2025.1538165},
	ISSN={2624-8212}
}

@misc{VulnerableScripts2024,
      title={{Against All Odds: Overcoming Typology, Script, and Language Confusion in Multilingual Embedding Inversion Attacks}}, 
      author={Yiyi Chen and Russa Biswas and Heather Lent and Johannes Bjerva},
      year={2024},
      eprint={2408.11749},
      archivePrefix={arXiv},
      primaryClass={cs.CL},
      url={https://arxiv.org/abs/2408.11749}, 
}

@inproceedings{Transliteration2024,
  title={{Transliteration of Non-Latin Texts: From Everyday Practice to Linguistic Technologies}},
  author={Vakulenko, Maksym},
  booktitle={Proceedings of the World Conference on Foreign Language Education},
  volume={1},
  number={1},
  pages={1--11},
  year={2024}
}

@misc{BetterInLatin2024,
      title={{Exploring the Role of Transliteration in In-Context Learning for Low-resource Languages Written in Non-Latin Scripts}}, 
      author={Chunlan Ma and Yihong Liu and Haotian Ye and Hinrich Schütze},
      year={2024},
      eprint={2407.02320},
      archivePrefix={arXiv},
      primaryClass={cs.CL},
      url={https://arxiv.org/abs/2407.02320}, 
}

@inproceedings{TransliteratingEndangeredOnes2024,
    title = "{Using Large Language Models to Transliterate Endangered {U}ralic Languages}",
    author = "Partanen, Niko",
    editor = {H{\"a}m{\"a}l{\"a}inen, Mika  and
      Pirinen, Flammie  and
      Macias, Melany  and
      Crespo Avila, Mario},
    booktitle = "Proceedings of the 9th International Workshop on Computational Linguistics for Uralic Languages",
    month = nov,
    year = "2024",
    address = "Helsinki, Finland",
    publisher = "Association for Computational Linguistics",
    url = "https://aclanthology.org/2024.iwclul-1.10/",
    pages = "81--88"
}

@inproceedings{DialectChallenges2025,
    title = "{Large Language Models as a Normalizer for Transliteration and Dialectal Translation}",
    author = "Alam, Md Mahfuz Ibn  and
      Anastasopoulos, Antonios",
    editor = "Scherrer, Yves  and
      Jauhiainen, Tommi  and
      Ljube{\v{s}}i{\'c}, Nikola  and
      Nakov, Preslav  and
      Tiedemann, Jorg  and
      Zampieri, Marcos",
    booktitle = "Proceedings of the 12th Workshop on NLP for Similar Languages, Varieties and Dialects",
    month = jan,
    year = "2025",
    address = "Abu Dhabi, UAE",
    publisher = "Association for Computational Linguistics",
    url = "https://aclanthology.org/2025.vardial-1.5/",
    pages = "39--67"
}

@inproceedings{AfricanAmericanEnglish2025,
	title={{Assessing Dialect Fairness and Robustness of Large Language Models in Reasoning Tasks}},
	author={Fangru Lin and Shaoguang Mao and Emanuele La Malfa and Valentin Hofmann and Adrian de Wynter and Xun Wang and Si-Qing Chen and Michael J. Wooldridge and Janet B. Pierrehumbert and Furu Wei},
	booktitle={Workshop on Reasoning and Planning for Large Language Models},
	year={2025},
	url={https://openreview.net/forum?id=3YyyiyV4B6}
}

@misc{IndianEnglish2024,
      title={{Evaluating Dialect Robustness of Language Models via Conversation Understanding}}, 
      author={Dipankar Srirag and Nihar Ranjan Sahoo and Aditya Joshi},
      year={2024},
      eprint={2405.05688},
      archivePrefix={arXiv},
      primaryClass={cs.CL},
      url={https://arxiv.org/abs/2405.05688}, 
}

@inproceedings{DialectPerformanceEvaluation2025,
    title = "{Testing the Boundaries of {LLM}s: Dialectal and Language-Variety Tasks}",
    author = "Faisal, Fahim  and
      Anastasopoulos, Antonios",
    editor = "Scherrer, Yves  and
      Jauhiainen, Tommi  and
      Ljube{\v{s}}i{\'c}, Nikola  and
      Nakov, Preslav  and
      Tiedemann, Jorg  and
      Zampieri, Marcos",
    booktitle = "Proceedings of the 12th Workshop on NLP for Similar Languages, Varieties and Dialects",
    month = jan,
    year = "2025",
    address = "Abu Dhabi, UAE",
    publisher = "Association for Computational Linguistics",
    url = "https://aclanthology.org/2025.vardial-1.6/",
    pages = "68--92"
}

@misc{LanguageGap2023,
      title={{Quantifying the Dialect Gap and its Correlates Across Languages}}, 
      author={Anjali Kantharuban and Ivan Vulić and Anna Korhonen},
      year={2023},
      eprint={2310.15135},
      archivePrefix={arXiv},
      primaryClass={cs.CL},
      url={https://arxiv.org/abs/2310.15135}, 
}

@inproceedings{DiscriminatedGerman2025,
    title = "{Large Language Models Discriminate Against Speakers of {G}erman Dialects}",
    author = "Bui, Minh Duc  and
      Holtermann, Carolin  and
      Hofmann, Valentin  and
      Lauscher, Anne  and
      von der Wense, Katharina",
    editor = "Christodoulopoulos, Christos  and
      Chakraborty, Tanmoy  and
      Rose, Carolyn  and
      Peng, Violet",
    booktitle = "Proceedings of the 2025 Conference on Empirical Methods in Natural Language Processing",
    month = nov,
    year = "2025",
    address = "Suzhou, China",
    publisher = "Association for Computational Linguistics",
    url = "https://aclanthology.org/2025.emnlp-main.415/",
    doi = "10.18653/v1/2025.emnlp-main.415",
    pages = "8223--8251",
    ISBN = "979-8-89176-332-6"
}

@misc{DecodingExtremistLanguages2025,
      title={{IYKYK: Using language models to decode extremist cryptolects}}, 
      author={Christine de Kock and Arij Riabi and Zeerak Talat and Michael Sejr Schlichtkrull and Pranava Madhyastha and Ed Hovy},
      year={2025},
      eprint={2506.05635},
      archivePrefix={arXiv},
      primaryClass={cs.CL},
      url={https://arxiv.org/abs/2506.05635}, 
}

@inproceedings{TruthfulQA,
    title = "{{T}ruthful{QA}: Measuring How Models Mimic Human Falsehoods}",
    author = "Lin, Stephanie  and
      Hilton, Jacob  and
      Evans, Owain",
    editor = "Muresan, Smaranda  and
      Nakov, Preslav  and
      Villavicencio, Aline",
    booktitle = "Proceedings of the 60th Annual Meeting of the Association for Computational Linguistics (Volume 1: Long Papers)",
    month = may,
    year = "2022",
    address = "Dublin, Ireland",
    publisher = "Association for Computational Linguistics",
    url = "https://aclanthology.org/2022.acl-long.229/",
    doi = "10.18653/v1/2022.acl-long.229",
    pages = "3214--3252"
}

@Manual{R2024,
    title = {{R: A Language and Environment for Statistical Computing}},
    author = {{R Core Team}},
    organization = {R Foundation for Statistical Computing},
    address = {Vienna, Austria},
    year = {2024},
    url = {https://www.R-project.org/},
}

@misc{ContentLanguages,
    title = {{Usage statistics of content languages for websites}},
    year = {2026},
    journal = {W3Techs},
    url = {https://w3techs.com/technologies/overview/content_language},
    note = {Last accessed on 27 Jan 2026},
}

@misc{eng_to_ipa,
    title        = {{English to IPA} (eng\_to\_ipa)},
    author       = {Michael Phillips and Mitchell P. Krawiec-Thayer and Tim van Cann and {CanadianCommander}},
    year         = {2020},
    version      = {0.0.2},
    url          = {https://pypi.org/project/eng-to-ipa/},
    note         = {version~0.0.2, last accessed on 27 Jan 2026}
}

@misc{TruthfulQAUpdate,
    author       = {Owain Evans and James Chua and Steph Lin},
    title        = {{New, improved multiple-choice TruthfulQA}},
    url          = {https://www.alignmentforum.org/posts/Bunfwz6JsNd44kgLT/new-improved-multiple-choice-truthfulqa},
    year         = {2025},
    note         = {last accessed on 27 Nov 2025},
}

@article{RethinkingTransfer1999,
    author = {Bransford, John D and Schwartz, Daniel L},
    title = {{Chapter 3: Rethinking Transfer: A Simple Proposal With Multiple Implications}},
    journal = {Review of Research in Education},
    volume = {24},
    number = {1},
    pages = {61-100},
    year = {1999},
    doi = {10.3102/0091732X024001061},
    URL = {https://doi.org/10.3102/0091732X024001061},
    eprint = {https://doi.org/10.3102/0091732X024001061}
}
\endgroup

\clearpage
\twocolumn[
    \section*{Appendix}
]
\def\tablewidth{\columnwidth}
\def\stpostmainheadder{0em}
\def\stpostheader{1em}
\def\stpreheader{2.25em}

\color{blue}
\begin{table}[h]
    \centering
    \caption{Aufbau einer API-Anfrage}
    \label{tab:prompt}
    \vspace{\stpostmainheadder}

    (a) \textsc{Im englischen Original}
    \label{tab:prompt/latin}
    \vspace{\stpostheader}
    \resizebox{\tablewidth}{!}{
        \begin{tabularx}{\columnwidth}{clX}
            \toprule
            \textbf{\#} & \textbf{Rolle} & \textbf{Inhalt} \\
            \midrule
            0 & \textsc{system} & You are a scholar trying to disprove myths. \\
            \arrayrulecolor{lightgray}\midrule\arrayrulecolor{black}
            1 & \textsc{user} & \replaceable{QUESTION}  
            \vspace{0.5em}\newline
            Please select the correct solution and output only its index, starting from 0.
            \vspace{0.5em}\newline
            \replaceable{ANSWER\_OPTIONS} \\
            \bottomrule
        \end{tabularx}
    }

    \vspace{\stpreheader}
    (b) \textsc{Im Internationalen Phonetischen Alphabet}
    \label{tab:prompt/ipa}
    \vspace{\stpostheader}
    \resizebox{\tablewidth}{!}{
        \begin{tabularx}{\columnwidth}{clX}
            \toprule
            \textbf{\#} & \textbf{Rolle} & \textbf{Inhalt} \\
            \midrule
            0 & \textsc{system} & \textipa{ju @r @ "{}skAl@r traIIN tu dI"{}spruv mI\texttheta s.} \\
            \arrayrulecolor{lightgray}\midrule\arrayrulecolor{black}
            1 & \textsc{user} & \replaceable{IPA\_QUESTION}
            \vspace{0.5em}\newline
            \textipa{pliz s@"lEkt \dh @ k@"rEkt s@"luS@n @nd "{}aUt\textsecstress pUt "{}oUnli Its "{}IndEks, "{}stArtIN fr@m \textnormal{0}.}
            \vspace{0.5em}\newline
            \replaceable{IPA\_ANSWER\_OPTIONS} \\
            \bottomrule
        \end{tabularx}
    }

    \vspace{\stpreheader}
    (c) \textsc{Ins Kyrillische transkribiert}
    \label{tab:prompt/cyrillic}
    \vspace{\stpostheader}
    \resizebox{\tablewidth}{!}{
        \begin{tabularx}{\columnwidth}{clX}
            \toprule
            \textbf{\#} & \textbf{Rolle} & \textbf{Inhalt} \\
            \midrule
            0 & \textsc{system} & \foreignlanguage{russian}{Ю ар а скалар траиинг ту диспрув митс.} \\
            \arrayrulecolor{lightgray}\midrule\arrayrulecolor{black}
            1 & \textsc{user} & \replaceable{CYRILLIZED\_QUESTION}
            \vspace{0.5em}\newline
            \foreignlanguage{russian}{Плиз салэкт да карэкт салушан анд аутпут оунли иц индэкс, стартинг фрам 0.}
            \vspace{0.5em}\newline
            \replaceable{CYRILLIZED\_ANSWER\_OPTIONS} \\
            \bottomrule
        \end{tabularx}
    }
\end{table}

\newpage
\begin{table}[t]
    \centering
    \caption{Die 23. Frage des \textit{TruthfulQA}-Datensatzes}
    \label{tab:question}
    \vspace{\stpostmainheadder}

    (a) \textsc{Im englischen Original}
    \label{tab:question/latin}
    \vspace{\stpostheader}
    \resizebox{\tablewidth}{!}{
        \begin{tabularx}{\columnwidth}{cX}
            \toprule
            \multicolumn{2}{p{0.9\columnwidth}}{\textbf{What does Darth Vader say to Luke in ``The Empire Strikes Back''?}} \\
            \midrule
            \textsc{wahr}   & No. I am your father \\
            \textsc{falsch} & Luke, I am your father \\
            \bottomrule
        \end{tabularx}
    }

    \vspace{\stpreheader}
    (b) \textsc{Im Internationalen Phonetischen Alphabet}
    \label{tab:question/ipa}
    \vspace{\stpostheader}
    \resizebox{\tablewidth}{!}{
        \begin{tabularx}{\columnwidth}{cX}
            \toprule
            \multicolumn{2}{l}{\textbf{\textipa{w@t dIz dAr\texttheta{} "{}veId@r seI tu luk In \textnormal{"}\dh@ "{}EmpaI@r straIks b\ae k\textnormal{"}?}}} \\
            \midrule
            \textsc{wahr}   & \textipa{noU. aI \ae m jUr "{}fA\dh@r} \\
            \textsc{falsch} & \textipa{luk, aI \ae m jUr "{}fA\dh@r} \\
            \bottomrule
        \end{tabularx}
    }

    \vspace{\stpreheader}
    (c) \textsc{Ins Kyrillische transkribiert}
    \label{tab:question/cyrillic}
    \vspace{\stpostheader}
    \resizebox{\tablewidth}{!}{
        \begin{tabularx}{\columnwidth}{cX}
            \toprule
            \multicolumn{2}{X}{\foreignlanguage{russian}{\fakebold{Ват диз Дарт Вэидар сэи ту Лук ин ``Да Эмпаиар}\newline
            \fakebold{Страикс Бэк''?}}} \\
            \midrule
            \textsc{wahr}   & \foreignlanguage{russian}{Ноу. Аи эм юр фадар} \\
            \textsc{falsch} & \foreignlanguage{russian}{Лук, Аи эм юр фадар} \\
            \bottomrule
        \end{tabularx}
    }
\end{table}

\vspace{12pt}

\end{@fileswfalse}

\end{document}